\documentclass{article} % For LaTeX2e
\usepackage{iclr2024_conference,times}

\usepackage{amsmath,amsfonts,bm}

\def\eqref#1{equation~\ref{#1}}
\def\1{\bm{1}}

\DeclareMathAlphabet{\mathsfit}{\encodingdefault}{\sfdefault}{m}{sl}
\SetMathAlphabet{\mathsfit}{bold}{\encodingdefault}{\sfdefault}{bx}{n}

\usepackage{hyperref}
\usepackage{url}

\usepackage{graphicx} 
\usepackage{multirow}

\title{CPLLM: Clinical Prediction with Large Language Models}

\author{Ofir Ben Shoham \\
Department of Software and \\Information Systems Engineering\\
Ben-Gurion University of the Negev\\
\texttt{benshoho@post.bgu.ac.il} \\
\And
Nadav Rappoport \\
Department of Software and \\Information Systems Engineering \\
Ben-Gurion University of the Negev \\
\texttt{nadavrap@bgu.ac.il} \\
}

\iclrfinalcopy % Uncomment for camera-ready version, but NOT for submission.
\begin{document}

\maketitle

\begin{abstract}
We present Clinical Prediction with Large Language Models (CPLLM), a method that involves fine-tuning a pre-trained Large Language Model (LLM) for clinical disease and readmission prediction. We utilized quantization and fine-tuned the LLM using prompts. For diagnosis prediction, we predict whether patients will be diagnosed with a target disease during their next visit or in the subsequent diagnosis, leveraging their historical diagnosis records. We compared our results to various baselines, including RETAIN, and Med-BERT, the current state-of-the-art model for disease prediction using temporal structured EHR data. In addition, We also evaluated CPLLM for patient hospital readmission prediction and compared our method's performance with benchmark baselines. Our experiments have shown that our proposed method, CPLLM, surpasses all the tested models in terms of PR-AUC and ROC-AUC metrics, showing state-of-the-art results for diagnosis prediction and patient hospital readmission prediction. Such a method can be easily implemented and integrated into the clinical process to help care providers estimate the next steps of patients.
\end{abstract}

\section{Introduction}
Large Language Models (LLMs) are a type of artificial intelligence (AI) that have been shown to be effective at a variety of Natural Language Processing tasks \citep{zhao2023survey}.
LLMs are trained on large amounts of textual data, which allows them to learn the statistical relationships between words and phrases. LLMs are used for different types of tasks, including natural language understanding, natural language generation, knowledge-intensive tasks, reasoning, and more \citep{yang2023harnessing}. This makes them well-suited for tasks that require understanding the meaning of a text, such as text classification \citep{gasparetto2022survey, sun2023text} and even clinical predictions in the medical domain \citep{thirunavukarasu2023large, steinberg2021language}.

Clinical predictions are used to estimate a patient's susceptibility to disease, gauge the likelihood of treatment response, or prognosticate the course of a patient's medical condition. 
\citep{laupacis1997clinical, wasson1985clinical}. These predictions have been executed via classical models such as Logistic Regression \citep{hosmer2013applied} and Random Forest \citep{breiman2001random}. However, these traditional methods do not model the order of the medical concept events (diagnoses, procedures, medications, etc.). Instead, they rely solely on the absence or presence of these events (features).

%Literature review. including the strengths and weaknesses of existing methods.
%A literature review was conducted to identify relevant studies %on the use of Deep-Learning models for clinical predictions. 
Modern event order prediction models, which are more advanced than the mentioned traditional prediction models, are based on RNNs or transformers, where the latter were shown to be superior \citep{vaswani2017attention}.
Specifically, BERT-Style Models like BERT \citep{devlin2018bert}, RoBERTa \citep{liu2019roberta}, and Deberta \citep{he2020deberta}. Another transformer-based architecture is GPT-style language model. GPT models are trained to generate the next word in a sequence. GPT models are used in a wide range of downstream tasks such as summarization, translation, question answering, and more \citep{floridi2020gpt}. To name a few GPT models: LLaMA \citep{touvron2023llama1, touvron2023llama2}, Falcon \citep{almazrouei2023falcon}, Bloom \citep{scao2022bloom}, and GPT4 \citep{OpenAI2023GPT4TR}. The flexibility and versatility of decoder-only models seem to be advantageous \citep{yang2023harnessing}.

The significance of the mentioned language models for handling sequential data is emphasized, particularly within the context of clinical prediction models relying on Electronic Health Record (EHR) data. Structured EHR data encompasses a patient's clinical history, notable for its irregular temporal sequence of events and observations \citep{steinberg2021language}. Previous works deal with modeling EHR diagnosis data as a sequence, such as BEHRT \citep{li2020behrt, li2022hi, shoham2023federated, meng2021bidirectional}, Med-BERT \citep{rasmy2021med} and Medic-BERT \citep{hansen2023patient} (for length of stay prediction), using BERT models. However, such BERT-based models represent each diagnosis code as an index and do not address the textual description of the ICD code. These models are pre-trained using clinical data, and have a limited sequence length input according to the BERT architecture.

There is limited research on using LLMs to train clinical prediction models. One of the main focus of applications of LLM in the clinic is on chat capability of these models \citep{singhal2023large, thirunavukarasu2023large} or using an LLM for medical texts-based tasks like text generation \citep{lu2022clinicalt5, agrawal2022large} and text comprehension \citep{yang2022large, sivarajkumar2022healthprompt, li2022clinical, jiang2023health}. In addition, \citet{chen-etal-2023-boosting} proposed a method called ClinTaT for cancer prediction. Their focus was on cancer prognostic prediction using few-shot learning, and their data modeling was not designed for structured EHR data that consists of a sequence of diagnoses. However, we want to harness the power of LLMs in understanding sequences of tokens derived from structured EHR data, specifically to train prediction models.
We represent the structured data as a text by representing each medical concept corresponding to a word, admissions are treated as visits, and patient history is considered a document.
% Objectives.
The objectives of this study are to develop a novel method for using LLMs to train clinical predictors and to evaluate the performance of this method on real-world datasets.

%Proposed method.
Our proposed method uses an LLM to predict future diagnoses and readmission of patients by fine-tuning LLMs. The medical concepts are represented by text descriptions. Fine-tuning is performed using a prompt that feeds the model with training samples. We used two different LLMs, Llama2, which is a general LLM \citep{touvron2023llama2} and BioMedLM, which was trained on biological and clinical text \citep{venigalla2022biomedlm}. We used four prediction tasks and two datasets and compared the performance to baseline models.

%Significance. % potential use with our method
The proposed method outperforms the state-of-the-art methods. Our generic method can be used for a variety of tasks and is not specific to any particular LLM. Moreover, our method is also suitable for different clinical domains such as demographics, diagnoses, laboratory test results, measurements, procedures, and more.

\textbf{Contributions}:
\textbf{(1)} We propose CPLLM, a novel method for clinical prediction with LLM that outperforms state-of-the-art models for disease prediction and patient readmission prediction for structured EHR data. In addition, CPLLM doesn't require pre-training on clinical data and achieves better performance than alternative approaches. Moreover, Our method has a longer sequence length limit compared to the baseline methods.
\textbf{(2)} We show that adding additional tokens to the pre-trained tokenizer of the LLM before fine-tuning improves the performance of the clinical prediction model.
\textbf{(3)} Our code is flexible for any LLM, available to use, and easily adaptable to various clinical prediction tasks.

\section{Methods}
\label{gen_inst}

\subsection{Disease prediction - problem definition}
Formally, for a given patient $p$, let $n$ denote the total number of diagnoses in their medical history. Thus, the patient's sequence of diagnoses is represented as $\{D_{p,1}, D_{p,2}, D_{p,3}, \ldots, D_{p,n}\}$, where each $D_{p,i}$ $(1 \leq i \leq n)$ corresponds to a medical diagnosis in the patient's history. We considered two types of diagnosis prediction: next diagnosis and next visit diagnosis.

\textbf{Next diagnosis prediction}: Predict whether patient $p$ will be diagnosed with a specific disease $D_x$ as the $D_{p,i+1}$ diagnosis given previous diagnoses. Our model relys on the patient's medical records up to the $i$-th diagnosis, denoted as $\{D_{p,1}, D_{p,2}, \ldots, D_{p,i}\}$. Where $D_{p,i}$ $(1 \leq i < n)$ indicates the most recent diagnosis observed for patient $p$. The predictive model utilizes this patient-specific historical medical information to determine whether patient $p$'s next diagnosis is a specific disease or not.

\textbf{Next visit diagnosis prediction}: Predicting the next diagnosis requires knowledge of the precise timing of each diagnosis. However, these data may occasionally be unavailable, such as when diagnoses are documented at the end of an admission. Consequently, in the context of the MIMIC-IV dataset, we undertake the task of forecasting whether a patient will receive a specific diagnosis in his subsequent admission.

\subsection{Patient Hospital readmission prediction}
Based on a patient's medical history, including procedures, diagnoses, and medications, our objective is to forecast whether the patient will experience hospital readmission within the next $X$ days. We follow the definition of $X$ as specified by the PyHealth benchmark \cite{pyhealth2023yang}. In our experiments with the MIMIC-IV dataset, we predict hospital readmission within a 15-day window, and for the eICU-CRD dataset, the prediction time-frame is 5 days (see section \ref{subsec:Data}).

% Please pay special attention to the instructions in section \ref{others}
% regarding figures, tables, acknowledgments, and references.

\subsection{Data}
\label{subsec:Data}
In this study, we used data from the eICU-CRD database \citep{pollard2018eicu} and data from the MIMIC-IV database \citep{johnson2020mimic}. Our datasets include ICD-9-CM (eICU-CRD) and ICD-10-CM (MIMIC-IV) diagnoses and their descriptions. In the eICU-CRD database, each diagnosis is associated with a timestamp. Consequently, we arranged the diagnoses in chronological order based on their respective diagnosis times. Our disease prediction task aims to anticipate whether the forthcoming diagnosis will correspond to a specific disease. Unlike the eICU-CRD dataset, the MIMIC-IV data lacks information on the exact time of each diagnosis assignment. However, it provides the start time for admission and the discharge times for each patient. As a result, our prediction task for this dataset revolves around determining whether a patient will be diagnosed with a specific disease during his subsequent visit.

Med-BERT adopts a pre-training strategy and trains BERT using Masked Language Modeling (MLM) and Length of stay (LOS) prediction tasks \citep{rasmy2021med}. Therefore, we extracted the necessary data from the databases, including the diagnosis codes for each patient. Additionally, we also include information on the LOS of each admission and the number of visits of each patient. On the other hand, in our approach, we did not conduct an additional pre-training step, as we focused on fine-tuning an LLM. In our proposed method, it's not required to note at which visit each diagnosis was given. Furthermore, the duration of hospital stay is not required. Notably, our method attains superior results even in the absence of these particulars. This aspect holds significance, since in certain situations, this data may not be accessible. For example, when a patient has not been admitted to the hospital but is under the care of a family doctor.

\textbf{Data Preprocessing}: For readmission prediction, we follow PyHealth's data preprocessing methodology. We include drugs, procedures, and diagnosis codes alongside their respective descriptions. Additionally, we incorporate both ICD-9 and ICD-10 codes and convert them to Clinical Classification Software (CCS) codes \cite{elixhauser2009clinical}. For drugs, we convert the codes to ATC codes \cite{nahler2009anatomical}. For procedures, we include ICD-9 and ICD-10 procedure codes and convert them to CCS codes using PyHealth.
For diagnosis prediction, for the MIMIC-IV dataset, we excluded patients with only one visit, as there is no medical history in such a case. Similarly, for the eICU-CRD dataset, patients with just one diagnosis were removed. We also excluded patients who have the disease we are trying to predict at the first visit (or the first diagnosis for eICU-CRD data).
We converted our ICD-10 codes to their corresponding CCS categories for MIMIC-IV, while for eICU-CRD, we retained the ICD-9 codes as they were. This decision was motivated by the higher number of ICD-10 codes compared to ICD-9 codes \citep{manchikanti2013ready}.
Based on the sequence of diagnoses for each patient, we determined whether the patient exhibited a specific diagnosis based on ICD diagnosis codes related to the specific disease according to the relevant CCS category \citep{elixhauser2014clinical}. 
Table \ref{tab:dataset_statistics} provides an overview of the number of patients, the count of final patients after preprocessing, average diagnoses, and average visits for each disease prediction task.

\subsubsection{Clinical Outcomes}
We evaluated our model for four prediction tasks: patient hospital readmission prediction and three diagnosis predictions covering Chronic kidney disease, Acute and unspecified renal failure, and Adult respiratory failure. The first two diagnoses were derived from the MIMIC-IV dataset, and the last was derived from the eICU-CRD dataset. The corresponding CCS codes for these diseases are 157 for Acute and unspecified renal failure, 158 for Chronic kidney disease, and 131 for Adult respiratory failure. For each prediction task, patients with specific disease ICD codes were assigned a positive label, and their diagnosis history encompassed all diagnostic codes recorded until the specific code indicated the outcome of interest.

\begin{table}[h]
\caption{Task statistics of the prediction tasks. Visit and diagnosis counts are calculated from the patient's medical history after preprocessing. IQR - Interquartile range.}
\label{tab:dataset_statistics}
\centering
\resizebox{\textwidth}{!}{
\begin{tabular}{|c|c|c|c|c|c|}
\hline
Dataset & Task & \# of patients & Final \# of patients & Median \# of visits (IQR) & Median \# of diagnoses (IQR) \\
\hline
MIMIC-IV & Chronic kidney disease & 84,453 & 26,161 & 1 (1-2) & 11 (7-19) \\
MIMIC-IV & Acute and unspecified renal failure & 84,453 & 26,736 & 1 (1-2) & 11 (7-19) \\
eICU-CRD & Adult respiratory failure & 132,677 & 56,419 & 1 (1-1) & 1 (1-2) \\

\hline
\end{tabular}
}
\end{table}

\subsection{Baseline Methods}
We conducted a rigorous performance assessment of the CPLLM against three baseline methods. For diagnosis prediction task, we used the next baseline models. First, Med-BERT with a classification layer \citep{rasmy2021med}. Second, with Logistic Regression \citep{hosmer2013applied}. Furthermore, we compared our method to RETAIN - a disease prediction model featuring double GRUs and attention modules \citep{choi2016retain}. We compared CPLLM with these baseline methods to gain valuable insights into its performance in clinical prediction downstream tasks. The comparison was conducted using two metrics: the area under the precision-recall curve (PR-AUC) and the area under the receiver operating characteristic curve (ROC-AUC). Disease prediction tasks are typically imbalanced; therefore ROC-AUC is less suitable for binary classifiers with imbalanced data \citet{davis2006relationship}. Therefore, our main evaluation metric is PR-AUC, but we also report ROC-AUC for consistency with the baseline methods. For readmission prediction, as mentioned earlier, we compared CPLLM with PyHealth baselines. The models we compared with include ConCare \cite{ma2020concare}, RETAIN \cite{choi2016retain}, deeper \cite{nguyen2016mathtt} and GRASP \cite{zhang2021grasp}.

\subsection{Our Proposed Method}
We propose a method called Clinical Prediction with Large Language Models (CPLLM). This method involves fine-tuning a LLM using prompts tailored to medical concept sequences. Through fine-tuning using prompts (inputs for LLM guidance), we direct the LLM to grasp intricate relationships among medical concepts. 

We utilized two LLMs: Llama2 (13B parameters) \citep{touvron2023llama2} and BioMedLM (also called PubMedGPT, 2.7B parameters) \citep{venigalla2022biomedlm}. To enhance the time and memory efficiency of fine-tuning these LLMs, we used QLoRA \citep{dettmers2023qlora} and PEFT \citep{houlsby2019parameter}. QLoRA is a PEFT approach that decreases the number of parameters requiring fine-tuning and also performs quantization \citep{dettmers2023qlora}. This combined approach effectively optimized the models' efficiency, enabling single-GPU fine-tuning for both BioMedLM and Llama2 models.

\begin{figure}[ht]
    \centering
    \includegraphics[width=1\textwidth]{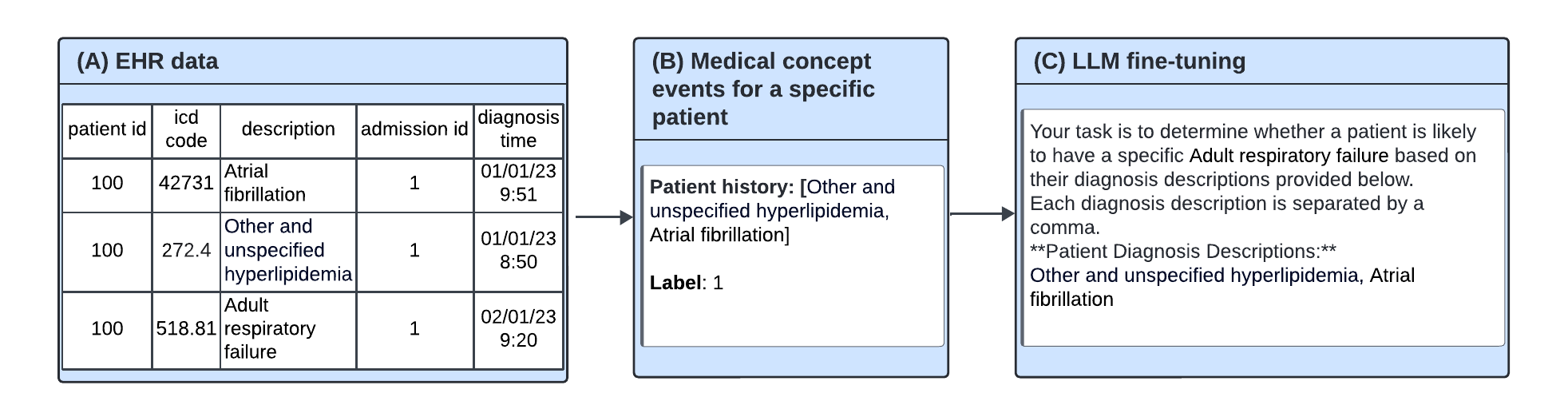}
    \caption{Illustration of the fine-tuning process for diagnosis prediction.
(A) An example of EHR structured data. The patient has three diagnoses.
(B) Patient's historical data is extracted from the EHR, and decoded to a textual list of descriptions.
(C) The decoded textual data is then injected into a designed prompt for fine-tuning the LLM. Fine-tuning prompts consist of a general description, the patient's diagnosis history, and a label. The label is set to 1 when the patient is diagnosed with the outcome of interest (e.g., Adult Respiratory Failure in the subsequent diagnosis or during the next admission, depending on the task.}
    \label{fig:CPLLM_training_flow}
\end{figure}

We performed separate fine-tuning of each LLM, leveraging specific prompts tailored to our patients' medical codes and their corresponding labels. In Figure \ref{fig:CPLLM_training_flow}, we present an example of the prompts utilized during the fine-tuning process for both the Llama2 and BioMedLM. We also indicated in the prompt the target disease, and the prompts were designed to incorporate the patients' individual medical code histories, with the goal of improving the models' performance. For readmission prediction, the prompt was very similar, but it included in addition drugs and procedures. For diagnosis prediction tasks, we added tokens of diagnosis descriptions missing from the original tokenizer vocabulary of the LLM. We performed an ablation study that compared the performance with and without changing the vocabulary of the pre-trained tokenizer.

For the clinical prediction downstream task, we performed fine-tuning as depicted in Figure \ref{fig:CPLLM_training_flow}. We used prompts to ask the LLMs to generate a single binary token in response (1 or 0) by adding a classification layer corresponding to the number of labels. By training the models with all patients' data for the specified number of epochs, we obtained the fine-tuned LLM tailored to our specific clinical prediction task.

\label{subsec:our_method}

\section{Experiments}
\subsection{Experimental Setup}
\label{sec:experiments_setup}
For readmission prediction, we compared our method to the PyHealth benchmark. For the diagnosis prediction tasks, We compare our method to three baseline models. The first is a simple Logistic Regression that does not model the data as a sequence but as simple independent, unordered variables \citep{manogaran2018health}. The second is RETAIN which is a two-level neural attention model \citep{choi2016retain}. The third baseline is Med-BERT, which is the state-of-the-art for structured EHR data for disease prediction. RETAIN was the baseline of Med-BERT. We split our data using an 70-10-20 ratio for train, validation, and train sets accordingly. For Med-BERT, we trained the pre-training model with the MLM and LOS tasks, with the TensorFlow package \citep{tensorflow2015-whitepaper}. The training of the Med-BERT's MLM phase was performed according to the fixed number of steps in the original implementation. The training took about 1.5 days on an RTX1080 GPU. Subsequently, we performed fine-tuning on the pre-trained model for the specific clinical prediction downstream tasks. The RETAIN and Med-BERT baselines trained for 500 epochs with early stopping based on the PR-AUC value derived from the validation set, using a maximum number of epochs without improvement of 5 epochs \citep{prechelt2002early}. During the training of the baselines, we experimented with various batch sizes \{32, 100\} and different learning rates \{\(1e^{-5}\), \(2e^{-5}\)\}. For each prediction task, we selected the hyper-parameters that achieved the best results on the validation set. For Logistic Regression training, we utilized the scikit-learn package \citep{pedregosa2011scikit} and trained the model on a CPU. To determine the optimal hyper-parameters for Logistic Regression, we conducted a grid search encompassing \textit{penalty} (L1 and L2 regularization), \textit{C}, \textit{solver}, and the maximum number of iterations. We explored values of \{0.1, 1, 10\} for \textit{C}, \{'liblinear', 'saga'\} for \textit{solver}, and \{100, 200, 500\} for the number of iterations. We took the best hyper-parameters based on the validation PR-AUC for each prediction task.

For CPLLM experiments, we fine-tuned two LLMs Llama2 (13B) and BioMedLM (2.7B) using HuggingFace \citep{wolf2019huggingface}. \citep{dettmers2023qlora}. Specifically, we used a learning rate of \(2e^{-5}\), Lora alpha of 32, Lora dropout of 0.1, and bias of none. Given the resource constraints, we meticulously determined and employed the maximum batch size that our GPU memory could accommodate. We fine-tuned each model over six epochs (and four epochs for readmission due to the larger dataset), selecting the best checkpoint based on validation PR-AUC. Fine-tuning Llama2 for six epochs required about a day of training on an RTX 6000 GPU, while BioMedLM took about two hours on the same hardware. Our fine-tuning process used PEFT, and we didn't perform additional pre-training in the clinical domain, yet our CPLLM method outperformed the baseline models.

\subsection{Results}
\subsubsection{Diagnosis Prediction Results}
We consider various models for the clinical prediction task: Logistic Regression, Med-BERT with a classification layer, RETAIN, and our proposed method called CPLLM. To examine the statistical significance of the results, we ran each model three times. Table \ref{tab:results_table} shows the mean and 95\% confidence interval of PR-AUC and ROC-AUC of these models.

Our findings demonstrate that our method, CPLLM, outperforms all tested models, including RETAIN, Med-BERT, and Logistic Regression, across both PR-AUC and ROC-AUC metrics. Specifically, in the context of the Adult respiratory failure task, CPLLM-Llama2 achieved a noteworthy PR-AUC value of 35.962\%, signifying an absolute improvement of 0.912\% over the best-performing baseline model, Logistic Regression, which obtained a PR-AUC score of 35.05\%. This improvement corresponds to a relative enhancement of 2.6\% in PR-AUC. Additionally, our method exhibits a relative increase of 5.1\% in PR-AUC when compared to RETAIN and a 3.31\% increase when compared to Med-BERT. Regarding ROC-AUC performance, CPLLM outperforms the baseline models. Furthermore, CPLLM-Llama2 demonstrates superior performance in this specific task compared to CPLLM-BioMedLM. Logistic Regression outperforms RETAIN in both PR-AUC (35.05\%) and ROC-AUC (74.664\%), but it also outperforms Med-BERT in PR-AUC, albeit not in ROC-AUC (74.664\% compared to 75.407\% for Med-BERT).

For Chronic kidney disease using the MIMIC-IV dataset, RETAIN had the worst performance in both metrics. Med-BERT outperformed Logistic Regression and RETAIN. CPLLM-Llama2 had the highest PR-AUC score of 33.992\%, followed by CPLLM-BioMedLM with 33.984\% and Med-BERT with 33.37\%. However, in ROC-AUC, CPLLM-BioMedLM outperformed all models with a score of 83.404\%, followed by CPLLM-Llama2 with 83.034\% and Med-BERT with 83.12\%.

For Acute and unspecified renal failure, CPLLM-Llama2 achieved the highest measurements, boasting a PR-AUC score of 45.442\% and an ROC-AUC score of 78.504\%. This signifies a notable improvement of 4.22\% in PR-AUC compared to the leading baseline model, RETAIN, in this task. Additionally, it demonstrates a 1.31\% improvement in ROC-AUC compared to the best-performing baseline, which is Logistic Regression with an ROC-AUC score of 77.486\%. Furthermore, it's worth highlighting that in this specific task, RETAIN outperforms Med-BERT in terms of PR-AUC but not ROC-AUC. Additionally, CPLLM-Llama2 demonstrates superior performance compared to CPLLM-BioMedLM. We found that CPLLM-Llama2 outperformed CPLLM-BioMedLM and therefore the rest of the analysis will be based on CPLLM-Llama2.

\begin{table}[ht]
    \caption{Performances of various models assessed across multiple tasks and datasets. The highest score per task is highlighted in bold.}
    \label{tab:results_table}
    \centering
    \begin{tabular}{|c|c|c|c|}
        \hline
        \textbf{Task} & \textbf{Model} & \textbf{PR-AUC} & \textbf{ROC-AUC} \\
        \hline
            \multirow{4}{*}{Adult respiratory failure} & Logistic Regression & 35.050 & 74.664 \\
        \cline{2-4}
        & RETAIN & 34.22 $\pm$ 0.299 & 74.454 $\pm$ 0.173 \\
        \cline{2-4}
        & Med-BERT & 34.81 $\pm$ 0.208 & 75.407 $\pm$ 0.073 \\
        \cline{2-4}
        & CPLLM-Llama2 & \textbf{35.962 $\pm$ 0.380} & \textbf{76.407 $\pm$ 0.262} \\
        \cline{2-4}
        & CPLLM-BioMedLM & 35.494 $\pm$ 0.352 & 75.975 $\pm$ 0.214 \\
        \hline

        \multirow{4}{*}{Chronic kidney disease} & Logistic Regression & 32.230 & 83.016 \\
        \cline{2-4}
        & RETAIN & 31.407 $\pm$ 1.379 & 81.692 $\pm$ 0.899 \\
        \cline{2-4}
        & Med-BERT & 33.37 $\pm$ 0.891 & 83.12 $\pm$ 0.173 \\
        \cline{2-4}
        & CPLLM-Llama2 & \textbf{33.992 $\pm$ 1.262} & 83.034 $\pm$ 0.511 \\
        \cline{2-4}
        & CPLLM-BioMedLM & 33.984 $\pm$ 1.077 & \textbf{83.404 $\pm$ 0.429} \\
        \hline

        \multirow{4}{*}{Acute and unspecified renal failure} & Logistic Regression & 42.075 & 77.486 \\
        \cline{2-4}
        & RETAIN & 43.603 $\pm$ 0.409 & 77.364 $\pm$ 0.394 \\
        \cline{2-4}
        & Med-BERT & 42.237 $\pm$ 0.408 & 77.427 $\pm$ 0.185 \\
        \cline{2-4}
        & CPLLM-Llama2 & \textbf{45.442 $\pm$ 0.839} & \textbf{78.504 $\pm$ 0.684} \\
        \cline{2-4}
        & CPLLM-BioMedLM & 45.161 $\pm$ 1.622 & 78.484 $\pm$ 0.403 \\
        \hline

    \end{tabular}
\end{table}

\subsubsection{Hospital Readmission Prediction Results}
To demonstrate the robustness of CPLLM, we expanded our analysis beyond diagnosis to include procedures and drugs. We compared CPLLM against several baseline methods from the PyHealth benchmark. Table \ref{tab:readmission} presents the results for patient hospital readmission prediction. In the case of MIMIC-IV, CPLLM with LLama2-13B achieved a PR-AUC of 68.986\%, outperforming ConCare, the second-best performing model, by 1.46\% (absolute). For eICU-CRD, CPLLM exhibited the highest PR-AUC among the baselines, achieving a PR-AUC of 94.115\%. Additionally, CPLLM achieved the highest ROC-AUC in both datasets.

\begin{table}[!ht]
\caption{PR-AUC and ROC-AUC performances of hospital readmission prediction task for MIMIC-IV and eICU-CRD datasets.}
\label{tab:readmission}
\centering
\small
\begin{tabular}{|c|c|c|c|}
\hline
\textbf{Dataset} & \textbf{Model} & \textbf{PR-AUC} & \textbf{ROC-AUC} \\
\hline
\multirow{5}{*}{MIMIC-IV} & CPLLM-Llama2 & \textbf{68.986 $\pm$ 0.499} & \textbf{68.155 $\pm$ 0.38} \\
\cline{2-4}
& ConCare & 67.523 $\pm$ 0.697 & 67.242 $\pm$ 0.269 \\
\cline{2-4}
& RETAIN & 67.343 $\pm$ 0.558 & 66.893 $\pm$ 0.421 \\
\cline{2-4}
& deeper & 66.891 $\pm$ 0.604 & 66.575 $\pm$ 0.371 \\
\cline{2-4}
& GRASP & 65.656 $\pm$ 2.929 & 65.302 $\pm$ 3.369 \\
\hline
\multirow{5}{*}{eICU-CRD} & CPLLM-Llama2 & \textbf{94.115 $\pm$ 0.704} & \textbf{77.916 $\pm$ 1.026} \\
\cline{2-4}
& ConCare & 93.429 $\pm$ 0.733 & 77.024 $\pm$ 1.156 \\
\cline{2-4}
& RETAIN & 93.615 $\pm$ 0.340 & 77.149 $\pm$ 1.048 \\
\cline{2-4}
& deeper & 93.814 $\pm$ 0.422 & 77.814 $\pm$ 0.385 \\
\cline{2-4}
& GRASP & 93.677 $\pm$ 1.824 & 77.515 $\pm$ 3.899 \\
\hline
\end{tabular}
\end{table}

\subsection{Ablation Study}
We conducted an ablation study to investigate the impact of the added tokens to the pre-trained tokenizer of the LLMs before fine-tuning. Table \ref{tab:added_tokens_ablation} provides a comprehensive overview of the PR-AUC and ROC-AUC, comparing scenarios with and without adding extra tokens.
For the task of predicting Acute and unspecified renal failure, adding the tokens yields enhancements in both PR-AUC and ROC-AUC for CPLLM-Llama2 (0.499\% absolute increase in PR-AUC and a 0.554\% absolute increase in ROC-AUC). Similarly, CPLLM-BioMedLM shows substantial improvements with a 1.631\% absolute increase in PR-AUC, representing a relative enhancement of 3.746\%, and a 0.414\% absolute increase in ROC-AUC.
In contrast, for the prediction of Chronic kidney disease, the inclusion of extra tokens does not significantly impact PR-AUC and ROC-AUC in the case of CPLLM-Llama2. However, CPLLM-BioMedLM demonstrates improvements, specifically an absolute enhancement of 0.686\% in ROC-AUC and an increase in PR-AUC from 32.638\% to 33.984\%. It is worth noting that the PR-AUC of BioMedLM exhibits less stability, as evidenced by a larger confidence interval when no additional tokens are employed (4.358\%). Nevertheless, we conducted two additional runs to get a better estimate of the PR-AUC. Subsequently, we observed that the PR-AUC for these five experiments amounted to 33.078\%, and the confidence intervals were reduced to 1.773\%.
For Adult respiratory failure prediction, the presence of additional tokens results in improved PR-AUC and ROC-AUC for CPLLM-Llama2, whereas it enhances PR-AUC but does not influence ROC-AUC for CPLLM-BioMedLM.
In summary, the findings of this ablation study suggest that, in the majority of cases (9 out of 12 measurements across three prediction tasks), incorporating the added tokens leads to enhanced performance in clinical prediction tasks.

\begin{table}[ht]
\centering
\begin{tabular}{|p{2.2cm}|p{3.0cm}|c|c|c|}
\hline
\multirow{2}{*}{Task} & \multirow{2}{*}{Model} & \multirow{2}{*}{Added Tokens} & \multirow{2}{*}{PR-AUC} & \multirow{2}{*}{ROC-AUC} \\ 
 &  &  &  & \\ \hline
\multirow{4}{*}{\shortstack[l]{Acute and \\unspecified\\ renal failure}} & \multirow{2}{*}{CPLLM-Llama2} & + & \textbf{45.442 $\pm$ 0.839} & \textbf{78.504 $\pm$ 0.684} \\ 
 & & - & 44.943 $\pm$ 1.268 & 77.95 $\pm$ 0.814 \\ 
\cline{2-5}

 & \multirow{2}{*}{CPLLM-BioMedLM} & + & \textbf{45.161 $\pm$ 1.622} & \textbf{78.484 $\pm$ 0.403} \\
 & & - & 43.53 $\pm$ 1.101 & 78.07 $\pm$ 0.625 
\\\hline

\multirow{4}{*}{\shortstack[l]{Chronic kidney\\ disease}} & \multirow{2}{*}{CPLLM-Llama2} & + & 33.992 $\pm$ 1.262 & 83.034 $\pm$ 0.511 \\ 
 & & - & \textbf{34.563 $\pm$ 1.578} & \textbf{83.178 $\pm$ 1.02} \\ 

\cline{2-5}
 & \multirow{2}{*}{CPLLM-BioMedLM} & + &\textbf{ 33.984 $\pm$ 1.077} &\textbf{ 83.404 $\pm$ 0.429} \\ 
 & & - & 32.638 $\pm$ 4.358 & 82.718 $\pm$ 1.191
\\\hline

\multirow{2}{*}{\shortstack[l]{\\Adult\\respiratory\\ failure}}  & \multirow{2}{*}{CPLLM-Llama2} & + & \textbf{35.962 $\pm$ 0.38} & \textbf{76.407 $\pm$ 0.262} \\ 
 & & - & 35.683 $\pm$ 0.164 & 75.776 $\pm$ 0.085 \\ 
\cline{2-5}

 & \multirow{2}{*}{CPLLM-BioMedLM} & + & 35.494 $\pm$ 0.352 & \textbf{75.975 $\pm$ 0.214} \\ 
 & & - & \textbf{35.714 $\pm$ 0.516} & 75.794 $\pm$ 0.194 \\ 
\hline

\end{tabular}
\caption{PR-AUC and ROC-AUC for CPLLM-Llama2 and CPLLM-BioMedLM, across three distinct medical tasks. Added Tokens column indicates whether additional tokens were incorporated into the pre-trained tokenizer. "+" and "-" - additional tokens were or were not added accordingly.}
\label{tab:added_tokens_ablation}
\end{table}

\section{Discussion}
% better performance, significant improvement. 
% we evaluated with icd10 and icd9 codes and two different datasets, eicu-crd that contains multi-center data and mimic. 
% CPLLM can be used for many clinical prediction taks. we showed for two tasks. 
Our proposed CPLLM method outperformed the baselines on all four tasks (3 diagnosis prediction and readmission prediction) across two different datasets. We used MIMIC-IV and eICU-CRD datasets to assess the model's ability to handle two diagnoses coding systems (ICD9 and ICD10) and two data types (homogeneous data from the same hospital in MIMIC-IV and multi-center data in eICU-CRD). CPLLM was superior to all baselines. CPLLM-Llama2 was the best model overall, and only for Chronic kidney disease did CPLLM-BioMedLM outperform CPLLM-Llama2, but only in terms of ROC-AUC. Using CPLLM-Llama2, we achieved PR-AUC relative improvements of 3.309\%, 1.864\%, and 7.588\% over Med-BERT on the three tasks, and ROC-AUC relative improvements of 1.326\% and 1.391\% on the Adult respiratory failure and Acute and unspecified renal failure prediction tasks. For hospital readmission prediction, CPLLM achieved relative improvements of 2.17\% compared to ConCare in PR-AUC for MIMIC-IV. For eICU-CRD readmission prediction, CPLLM showed a relative improvement of 0.31\% compared to the second-best result, deeper.

% no need for another pre-training.
Unlike existing approaches that necessitate pre-training with medical concept sequences, our method eliminates the need for additional pre-training tasks. For instance, Med-BERT entails both MLM and LOS prediction tasks using patient sequences of medical concepts. Based on our findings and results, it's evident that LLMs possess the capability to adeptly represent sequential clinical data without the need for specific pre-training based on clinical sequences. Beyond that, our method can be used even without the LOS data of each patient's hospitalizations, which is required for Med-BERT pre-training. Sometimes, these data are not available, for example, when there is no hospitalization, but rather data collected among patients who visited a physician in outpatient settings, or when LOS data is not available like in claims data. 
% no visits usage.
Furthermore, during the fine-tuning training of CPLLM, it is not necessary to know which diagnoses were given in which visit but only the diagnoses as a sequence. This differs from Med-BERT, which relies on this information for fine-tuning. Notably, we achieved superior performance even without these specific details.

% the additional tokens improves the measurements.
We found that including additional tokens in the LLM's tokenizer before fine-tuning improves the measurement of the prediction model in most cases. For instance, as Llama2 was not initially pre-trained on clinical data, supplementing it with missing description codes can enhance its understanding of the medical domain.

% retain vs Med-BERT.
% ROC-AUC vs PR-AUC. 
% Med-BERT and retain vs Logistic Regression. 
In the original Med-BERT paper, improvements over RETAIN were demonstrated in terms of ROC-AUC for three disease prediction tasks \citep{rasmy2021med}. We also found that Med-BERT consistently outperformed RETAIN in all prediction tasks based on ROC-AUC. However, it's worth noting that, as previously mentioned, ROC-AUC may not be an optimal metric for imbalanced datasets \citep{davis2006relationship}. In contrast, when considering PR-AUC, Med-BERT exhibited superior performance compared to RETAIN in two out of three tasks, although it did not outperform RETAIN in the prediction of Acute and unspecified renal failure (with PR-AUC values of 43.603\% for RETAIN and 42.237\% for Med-BERT), despite achieving a higher ROC-AUC than RETAIN.

% our method is generic to any base model and any medical concepts.
In our readmission prediction experiment, which included diagnoses, drugs, and procedures, we demonstrated the flexibility of our method. It seamlessly incorporates medical concepts from various domains into the sequence with minimal adjustments to the prompt text.

% training time of Med-BERT vs CPLLM-BioMedLM. 

% Accountability, Fairness, data privacy and selection, transparency, explainability, value and purpose alignment {Attention is not all you need: the complicated case of ethically using large language models in healthcare and medicine, https://doi.org/10.1016/j.ebiom.2023.104512}

% our method can be useful for long text, compared to Med-BERT and RETAIN which are limited. 
Another strength of our proposed method lies in its remarkable capacity to handle longer sequences compared to the current state-of-the-art models for structured EHR data. With maximum sequence lengths of 1024 tokens for CPLLM-BioMedLM and 4096 tokens for CPLLM-Llama2, our approach far surpasses the limitations imposed by Med-BERT and BEHRT \citep{li2020behrt}. Both Med-BERT and BEHRT are constrained by BERT's maximum of 512 tokens, which significantly restricts their ability to handle longer inputs \citep{devlin2018bert}. Without the need for additional training, our method also handles longer sequences compared to Hi-BEHRT, which is specially trained and designed to handle sequences with a maximum of 1220 tokens \citep{li2022hi}.

Healthcare stakeholders are looking for ways to improve caregiving without risking patient's data. The two LLMs we tested are such that can be deployed and used on-premise or in a secure environment and do not require sharing personal data over the web.
% advantages: simple to change our base model. 
% Limitations:
% - Ethical and privacy when using healthcare and medicine data.
% - Fine-tuning time. You wrote "Fine-tuning Llama2 for six epochs required about one day of training on an RTX 6000 GPU, while BioMedLM took about two hours on the same hardware."
% - inference time
% long sequences 
% inference time
% storage
% prompt is required here.

%\textbf{Limitations}:
While our method demonstrates promising results in utilizing LLMs for clinical prediction tasks, it is important to acknowledge several limitations.
While our method accommodates sequences of up to 4096 tokens for CPLLM-Llama2 and 1024 tokens for CPLLM-BioMedLM, our tests did not include exceptionally long sequences that could fully explore the implications of this extended token limit. That is because the datasets we used do not contain very long observations or many diagnoses of a single patient. Moreover, due to the greater number of parameters in LLMs, our method demands more computational resources, inference time, and training time. Specifically, CPLLM-Llama2 had a longer training time than Med-BERT. However, CPLLM-BioMedLM requires less training time compared to Med-BERT (\ref{sec:experiments_setup}). That's because CPLLM-BioMedLM does not require additional pre-training, unlike necessity for MLM and LOS pre-training in Med-BERT.

In addition, in our method, there is a necessity to use a specific prompt, a requirement that does not apply to the baseline models. As a result, sometimes the prompt needs to be adapted according to a base model.

% future work: combine RAG, chain of thought. 
\textbf{Future work}: We hypothesize that combining a retrieval augmentation \citep{mialon2023augmented, zakka2024almanac}, with LLM can improve performance. This is because it allows to include general updated knowledge about the diseases that the patient has been diagnosed with in their medical history. Additionally, this approach can incorporate general knowledge and known risk factors into research on the disease we are trying to predict.

\section{Conclusion}
In this work, we presented CPLLM, a novel method for clinical disease prediction and patient hospital readmission prediction based on the clinical history of patients. CPLLM has practical application potential. By surpassing the state-of-the-art in clinical task prediction, our method enables more accurate and robust disease forecasting, as well as patient hospital readmission forecasting. CPLLM demonstrated superior performance across all three four on two datasets (MIMIC-IV and eICU-CRD). It processes ICD9 and ICD10 diagnoses, procedures, and drugs. We showcased its robustness in dealing with homogeneous and multi-center data. Our method's advantage lies in eliminating the need for additional pre-training tasks, unlike Med-BERT. Furthermore, our method remains adaptable the length of stay data is unavailable, making it suitable for a broader range of healthcare scenarios, including those involving non-hospitalized patients. In addition, CPLLM's fine-tuning process requires patients' diagnoses as a sequence, without the need for which diagnoses were given in which visit. Notably, our method can handle much longer sequences than existing state-of-the-art models.

\section{Reproducibility}
% Our code is provided in the Supplementary Material. 
Our code is available at the following link: \href{https://github.com/nadavlab/CPLLM}{https://github.com/nadavlab/CPLLM}.
Implementation details can be found in the Experimental Setup section \ref{sec:experiments_setup}. To execute the baseline code, we used the source code published as part of the Med-BERT paper \citep{rasmy2021med}.

For our experiments, we used the MIMIC-IV v2.0 dataset \citep{johnson2020mimic}, accessible at \href{https://physionet.org/content/mimiciv/2.0/}{https://physionet.org/content/mimiciv/2.0/}, as well as the eICU-CRD multi-center dataset \citep{pollard2018eicu}, which can be found at \href{https://physionet.org/content/eicu-crd/2.0/}{https://physionet.org/content/eicu-crd/2.0/}.

\bibliography{iclr2024_conference}

\begin{thebibliography}{56}
\providecommand{\natexlab}[1]{#1}
\providecommand{\url}[1]{\texttt{#1}}
\expandafter\ifx\csname urlstyle\endcsname\relax
  \providecommand{\doi}[1]{doi: #1}\else
  \providecommand{\doi}{doi: \begingroup \urlstyle{rm}\Url}\fi

\bibitem[Abadi et~al.(2015)Abadi, Agarwal, Barham, Brevdo, Chen, Citro, Corrado, Davis, Dean, Devin, Ghemawat, Goodfellow, Harp, Irving, Isard, Jia, Jozefowicz, Kaiser, Kudlur, Levenberg, Man\'{e}, Monga, Moore, Murray, Olah, Schuster, Shlens, Steiner, Sutskever, Talwar, Tucker, Vanhoucke, Vasudevan, Vi\'{e}gas, Vinyals, Warden, Wattenberg, Wicke, Yu, and Zheng]{tensorflow2015-whitepaper}
Mart\'{i}n Abadi, Ashish Agarwal, Paul Barham, Eugene Brevdo, Zhifeng Chen, Craig Citro, Greg~S. Corrado, Andy Davis, Jeffrey Dean, Matthieu Devin, Sanjay Ghemawat, Ian Goodfellow, Andrew Harp, Geoffrey Irving, Michael Isard, Yangqing Jia, Rafal Jozefowicz, Lukasz Kaiser, Manjunath Kudlur, Josh Levenberg, Dandelion Man\'{e}, Rajat Monga, Sherry Moore, Derek Murray, Chris Olah, Mike Schuster, Jonathon Shlens, Benoit Steiner, Ilya Sutskever, Kunal Talwar, Paul Tucker, Vincent Vanhoucke, Vijay Vasudevan, Fernanda Vi\'{e}gas, Oriol Vinyals, Pete Warden, Martin Wattenberg, Martin Wicke, Yuan Yu, and Xiaoqiang Zheng.
\newblock {TensorFlow}: Large-scale machine learning on heterogeneous systems, 2015.
\newblock URL \url{https://www.tensorflow.org/}.
\newblock Software available from tensorflow.org.

\bibitem[Agrawal et~al.(2022)Agrawal, Hegselmann, Lang, Kim, and Sontag]{agrawal2022large}
Monica Agrawal, Stefan Hegselmann, Hunter Lang, Yoon Kim, and David Sontag.
\newblock Large language models are few-shot clinical information extractors.
\newblock In \emph{Proceedings of the 2022 Conference on Empirical Methods in Natural Language Processing}, pp.\  1998--2022, 2022.

\bibitem[Almazrouei et~al.(2023)Almazrouei, Alobeidli, Alshamsi, Cappelli, Cojocaru, Debbah, Goffinet, Heslow, Launay, Malartic, et~al.]{almazrouei2023falcon}
Ebtesam Almazrouei, Hamza Alobeidli, Abdulaziz Alshamsi, Alessandro Cappelli, Ruxandra Cojocaru, Merouane Debbah, Etienne Goffinet, Daniel Heslow, Julien Launay, Quentin Malartic, et~al.
\newblock Falcon-40b: an open large language model with state-of-the-art performance.
\newblock \emph{Findings of the Association for Computational Linguistics: ACL}, 2023:\penalty0 10755--10773, 2023.

\bibitem[Breiman(2001)]{breiman2001random}
Leo Breiman.
\newblock Random forests.
\newblock \emph{Machine learning}, 45:\penalty0 5--32, 2001.

\bibitem[Chen et~al.(2023)Chen, Micsinai~Balan, and Brown]{chen-etal-2023-boosting}
Zekai Chen, Mariann Micsinai~Balan, and Kevin Brown.
\newblock Boosting transformers and language models for clinical prediction in immunotherapy.
\newblock In \emph{Proceedings of the 61st Annual Meeting of the Association for Computational Linguistics (Volume 5: Industry Track)}, pp.\  332--340, Toronto, Canada, July 2023. Association for Computational Linguistics.
\newblock \doi{10.18653/v1/2023.acl-industry.32}.
\newblock URL \url{https://aclanthology.org/2023.acl-industry.32}.

\bibitem[Choi et~al.(2016)Choi, Bahadori, Sun, Kulas, Schuetz, and Stewart]{choi2016retain}
Edward Choi, Mohammad~Taha Bahadori, Jimeng Sun, Joshua Kulas, Andy Schuetz, and Walter Stewart.
\newblock Retain: An interpretable predictive model for healthcare using reverse time attention mechanism.
\newblock \emph{Advances in neural information processing systems}, 29, 2016.

\bibitem[Davis \& Goadrich(2006)Davis and Goadrich]{davis2006relationship}
Jesse Davis and Mark Goadrich.
\newblock The relationship between precision-recall and roc curves.
\newblock In \emph{Proceedings of the 23rd international conference on Machine learning}, pp.\  233--240, 2006.

\bibitem[Dettmers et~al.(2023)Dettmers, Pagnoni, Holtzman, and Zettlemoyer]{dettmers2023qlora}
Tim Dettmers, Artidoro Pagnoni, Ari Holtzman, and Luke Zettlemoyer.
\newblock Qlora: Efficient finetuning of quantized llms.
\newblock \emph{arXiv preprint arXiv:2305.14314}, 2023.

\bibitem[Devlin et~al.(2018)Devlin, Chang, Lee, and Toutanova]{devlin2018bert}
Jacob Devlin, Ming-Wei Chang, Kenton Lee, and Kristina Toutanova.
\newblock Bert: Pre-training of deep bidirectional transformers for language understanding.
\newblock \emph{arXiv preprint arXiv:1810.04805}, 2018.

\bibitem[Elixhauser(2009)]{elixhauser2009clinical}
Anne Elixhauser.
\newblock Clinical classifications software (ccs) 2009.
\newblock \emph{http://www. hcug-us. ahrq. gov/toolssoft-ware/ccs/ccs. jsp}, 2009.

\bibitem[Elixhauser et~al.(2014)Elixhauser, Steiner, and Palmer]{elixhauser2014clinical}
Anne Elixhauser, Claudia Steiner, and L~Palmer.
\newblock Clinical classifications software (ccs).
\newblock \emph{US agency for healthcare research and quality}, 2014, 2014.

\bibitem[Floridi \& Chiriatti(2020)Floridi and Chiriatti]{floridi2020gpt}
Luciano Floridi and Massimo Chiriatti.
\newblock Gpt-3: Its nature, scope, limits, and consequences.
\newblock \emph{Minds and Machines}, 30:\penalty0 681--694, 2020.

\bibitem[Gasparetto et~al.(2022)Gasparetto, Marcuzzo, Zangari, and Albarelli]{gasparetto2022survey}
Andrea Gasparetto, Matteo Marcuzzo, Alessandro Zangari, and Andrea Albarelli.
\newblock A survey on text classification algorithms: From text to predictions.
\newblock \emph{Information}, 13\penalty0 (2):\penalty0 83, 2022.

\bibitem[Hansen et~al.(2023)Hansen, Nielsen, Mulvad, Strausholm, Sagi, and Hose]{hansen2023patient}
Emil~Riis Hansen, Thomas~Dyhre Nielsen, Thomas Mulvad, Mads~Nibe Strausholm, Tomer Sagi, and Katja Hose.
\newblock Patient event sequences for predicting hospitalization length of stay.
\newblock In \emph{International Conference on Artificial Intelligence in Medicine}, pp.\  51--56. Springer, 2023.

\bibitem[He et~al.(2020)He, Liu, Gao, and Chen]{he2020deberta}
Pengcheng He, Xiaodong Liu, Jianfeng Gao, and Weizhu Chen.
\newblock Deberta: Decoding-enhanced bert with disentangled attention.
\newblock \emph{arXiv preprint arXiv:2006.03654}, 2020.

\bibitem[Hosmer~Jr et~al.(2013)Hosmer~Jr, Lemeshow, and Sturdivant]{hosmer2013applied}
David~W Hosmer~Jr, Stanley Lemeshow, and Rodney~X Sturdivant.
\newblock \emph{Applied logistic regression}, volume 398.
\newblock John Wiley \& Sons, 2013.

\bibitem[Houlsby et~al.(2019)Houlsby, Giurgiu, Jastrzebski, Morrone, De~Laroussilhe, Gesmundo, Attariyan, and Gelly]{houlsby2019parameter}
Neil Houlsby, Andrei Giurgiu, Stanislaw Jastrzebski, Bruna Morrone, Quentin De~Laroussilhe, Andrea Gesmundo, Mona Attariyan, and Sylvain Gelly.
\newblock Parameter-efficient transfer learning for nlp.
\newblock In \emph{International Conference on Machine Learning}, pp.\  2790--2799. PMLR, 2019.

\bibitem[Jiang et~al.(2023)Jiang, Liu, Nejatian, Nasir-Moin, Wang, Abidin, Eaton, Riina, Laufer, Punjabi, et~al.]{jiang2023health}
Lavender~Yao Jiang, Xujin~Chris Liu, Nima~Pour Nejatian, Mustafa Nasir-Moin, Duo Wang, Anas Abidin, Kevin Eaton, Howard~Antony Riina, Ilya Laufer, Paawan Punjabi, et~al.
\newblock Health system-scale language models are all-purpose prediction engines.
\newblock \emph{Nature}, pp.\  1--6, 2023.

\bibitem[Johnson et~al.(2020)Johnson, Bulgarelli, Pollard, Horng, Celi, and Mark]{johnson2020mimic}
Alistair Johnson, Lucas Bulgarelli, Tom Pollard, Steven Horng, Leo~Anthony Celi, and Roger Mark.
\newblock Mimic-iv.
\newblock \emph{PhysioNet. Available online at: https://physionet. org/content/mimiciv/1.0/(accessed August 23, 2021)}, 2020.

\bibitem[Laupacis et~al.(1997)Laupacis, Sekar, et~al.]{laupacis1997clinical}
Andreas Laupacis, Nandita Sekar, et~al.
\newblock Clinical prediction rules: a review and suggested modifications of methodological standards.
\newblock \emph{Jama}, 277\penalty0 (6):\penalty0 488--494, 1997.

\bibitem[Li et~al.(2020)Li, Rao, Solares, Hassaine, Ramakrishnan, Canoy, Zhu, Rahimi, and Salimi-Khorshidi]{li2020behrt}
Yikuan Li, Shishir Rao, Jos{\'e} Roberto~Ayala Solares, Abdelaali Hassaine, Rema Ramakrishnan, Dexter Canoy, Yajie Zhu, Kazem Rahimi, and Gholamreza Salimi-Khorshidi.
\newblock Behrt: transformer for electronic health records.
\newblock \emph{Scientific reports}, 10\penalty0 (1):\penalty0 7155, 2020.

\bibitem[Li et~al.(2022{\natexlab{a}})Li, Mamouei, Salimi-Khorshidi, Rao, Hassaine, Canoy, Lukasiewicz, and Rahimi]{li2022hi}
Yikuan Li, Mohammad Mamouei, Gholamreza Salimi-Khorshidi, Shishir Rao, Abdelaali Hassaine, Dexter Canoy, Thomas Lukasiewicz, and Kazem Rahimi.
\newblock Hi-behrt: Hierarchical transformer-based model for accurate prediction of clinical events using multimodal longitudinal electronic health records.
\newblock \emph{IEEE journal of biomedical and health informatics}, 27\penalty0 (2):\penalty0 1106--1117, 2022{\natexlab{a}}.

\bibitem[Li et~al.(2022{\natexlab{b}})Li, Wehbe, Ahmad, Wang, and Luo]{li2022clinical}
Yikuan Li, Ramsey~M Wehbe, Faraz~S Ahmad, Hanyin Wang, and Yuan Luo.
\newblock Clinical-longformer and clinical-bigbird: Transformers for long clinical sequences.
\newblock \emph{arXiv preprint arXiv:2201.11838}, 2022{\natexlab{b}}.

\bibitem[Liu et~al.(2019)Liu, Ott, Goyal, Du, Joshi, Chen, Levy, Lewis, Zettlemoyer, and Stoyanov]{liu2019roberta}
Yinhan Liu, Myle Ott, Naman Goyal, Jingfei Du, Mandar Joshi, Danqi Chen, Omer Levy, Mike Lewis, Luke Zettlemoyer, and Veselin Stoyanov.
\newblock Roberta: A robustly optimized bert pretraining approach.
\newblock \emph{arXiv preprint arXiv:1907.11692}, 2019.

\bibitem[Lu et~al.(2022)Lu, Dou, and Nguyen]{lu2022clinicalt5}
Qiuhao Lu, Dejing Dou, and Thien Nguyen.
\newblock Clinicalt5: A generative language model for clinical text.
\newblock In \emph{Findings of the Association for Computational Linguistics: EMNLP 2022}, pp.\  5436--5443, 2022.

\bibitem[Ma et~al.(2020)Ma, Zhang, Wang, Ruan, Wang, Tang, Ma, Gao, and Gao]{ma2020concare}
Liantao Ma, Chaohe Zhang, Yasha Wang, Wenjie Ruan, Jiangtao Wang, Wen Tang, Xinyu Ma, Xin Gao, and Junyi Gao.
\newblock Concare: Personalized clinical feature embedding via capturing the healthcare context.
\newblock In \emph{Proceedings of the AAAI Conference on Artificial Intelligence}, volume~34, pp.\  833--840, 2020.

\bibitem[Manchikanti et~al.(2013)Manchikanti, Falco, and Hirsch]{manchikanti2013ready}
Laxmaiah Manchikanti, Frank~JE Falco, and Joshua~A Hirsch.
\newblock Ready or not! here comes icd-10.
\newblock \emph{Journal of neurointerventional surgery}, 5\penalty0 (1):\penalty0 86--91, 2013.

\bibitem[Manogaran \& Lopez(2018)Manogaran and Lopez]{manogaran2018health}
Gunasekaran Manogaran and Daphne Lopez.
\newblock Health data analytics using scalable logistic regression with stochastic gradient descent.
\newblock \emph{International Journal of Advanced Intelligence Paradigms}, 10\penalty0 (1-2):\penalty0 118--132, 2018.

\bibitem[Meng et~al.(2021)Meng, Speier, Ong, and Arnold]{meng2021bidirectional}
Yiwen Meng, William Speier, Michael~K Ong, and Corey~W Arnold.
\newblock Bidirectional representation learning from transformers using multimodal electronic health record data to predict depression.
\newblock \emph{IEEE Journal of Biomedical and Health Informatics}, 25\penalty0 (8):\penalty0 3121--3129, 2021.

\bibitem[Mialon et~al.(2023)Mialon, Dess{\`i}, Lomeli, Nalmpantis, Pasunuru, Raileanu, Rozi{\`e}re, Schick, Dwivedi-Yu, Celikyilmaz, Grave, LeCun, and Scialom]{mialon2023augmented}
Gr{\'e}goire Mialon, Roberto Dess{\`i}, Maria Lomeli, Christoforos Nalmpantis, Ramakanth Pasunuru, Roberta Raileanu, Baptiste Rozi{\`e}re, Timo Schick, Jane Dwivedi-Yu, Asli Celikyilmaz, Edouard Grave, Yann LeCun, and Thomas Scialom.
\newblock Augmented language models: a survey.
\newblock \emph{ArXiv}, 2023.

\bibitem[Nahler \& Nahler(2009)Nahler and Nahler]{nahler2009anatomical}
Gerhard Nahler and Gerhard Nahler.
\newblock Anatomical therapeutic chemical classification system (atc).
\newblock \emph{Dictionary of Pharmaceutical Medicine}, pp.\  8--8, 2009.

\bibitem[Nguyen et~al.(2016)Nguyen, Tran, Wickramasinghe, and Venkatesh]{nguyen2016mathtt}
Phuoc Nguyen, Truyen Tran, Nilmini Wickramasinghe, and Svetha Venkatesh.
\newblock Deepr: a convolutional net for medical records.
\newblock \emph{IEEE journal of biomedical and health informatics}, 21\penalty0 (1):\penalty0 22--30, 2016.

\bibitem[OpenAI(2023)]{OpenAI2023GPT4TR}
OpenAI.
\newblock Gpt-4 technical report.
\newblock \emph{arXiv preprint arXiv:2303.08774}, 2023.

\bibitem[Pedregosa et~al.(2011)Pedregosa, Varoquaux, Gramfort, Michel, Thirion, Grisel, Blondel, Prettenhofer, Weiss, Dubourg, et~al.]{pedregosa2011scikit}
Fabian Pedregosa, Ga{\"e}l Varoquaux, Alexandre Gramfort, Vincent Michel, Bertrand Thirion, Olivier Grisel, Mathieu Blondel, Peter Prettenhofer, Ron Weiss, Vincent Dubourg, et~al.
\newblock Scikit-learn: Machine learning in python.
\newblock \emph{the Journal of machine Learning research}, 12:\penalty0 2825--2830, 2011.

\bibitem[Pollard et~al.(2018)Pollard, Johnson, Raffa, Celi, Mark, and Badawi]{pollard2018eicu}
Tom~J Pollard, Alistair~EW Johnson, Jesse~D Raffa, Leo~A Celi, Roger~G Mark, and Omar Badawi.
\newblock The eicu collaborative research database, a freely available multi-center database for critical care research.
\newblock \emph{Scientific data}, 5\penalty0 (1):\penalty0 1--13, 2018.

\bibitem[Prechelt(2002)]{prechelt2002early}
Lutz Prechelt.
\newblock Early stopping-but when?
\newblock In \emph{Neural Networks: Tricks of the trade}, pp.\  55--69. Springer, 2002.

\bibitem[Rasmy et~al.(2021)Rasmy, Xiang, Xie, Tao, and Zhi]{rasmy2021med}
Laila Rasmy, Yang Xiang, Ziqian Xie, Cui Tao, and Degui Zhi.
\newblock Med-bert: pretrained contextualized embeddings on large-scale structured electronic health records for disease prediction.
\newblock \emph{NPJ digital medicine}, 4\penalty0 (1):\penalty0 86, 2021.

\bibitem[Scao et~al.(2022)Scao, Fan, Akiki, Pavlick, Ili{\'c}, Hesslow, Castagn{\'e}, Luccioni, Yvon, Gall{\'e}, et~al.]{scao2022bloom}
Teven~Le Scao, Angela Fan, Christopher Akiki, Ellie Pavlick, Suzana Ili{\'c}, Daniel Hesslow, Roman Castagn{\'e}, Alexandra~Sasha Luccioni, Fran{\c{c}}ois Yvon, Matthias Gall{\'e}, et~al.
\newblock Bloom: A 176b-parameter open-access multilingual language model.
\newblock \emph{arXiv preprint arXiv:2211.05100}, 2022.

\bibitem[Shoham \& Rappoport(2023)Shoham and Rappoport]{shoham2023federated}
Ofir~Ben Shoham and Nadav Rappoport.
\newblock Federated learning of medical concepts embedding using behrt.
\newblock \emph{arXiv preprint arXiv:2305.13052}, 2023.

\bibitem[Singhal et~al.(2023)Singhal, Azizi, Tu, Mahdavi, Wei, Chung, Scales, Tanwani, Cole-Lewis, Pfohl, et~al.]{singhal2023large}
Karan Singhal, Shekoofeh Azizi, Tao Tu, S~Sara Mahdavi, Jason Wei, Hyung~Won Chung, Nathan Scales, Ajay Tanwani, Heather Cole-Lewis, Stephen Pfohl, et~al.
\newblock Large language models encode clinical knowledge.
\newblock \emph{Nature}, pp.\  1--9, 2023.

\bibitem[Sivarajkumar \& Wang(2022)Sivarajkumar and Wang]{sivarajkumar2022healthprompt}
Sonish Sivarajkumar and Yanshan Wang.
\newblock Healthprompt: A zero-shot learning paradigm for clinical natural language processing.
\newblock In \emph{AMIA Annual Symposium Proceedings}, volume 2022, pp.\  972. American Medical Informatics Association, 2022.

\bibitem[Steinberg et~al.(2021)Steinberg, Jung, Fries, Corbin, Pfohl, and Shah]{steinberg2021language}
Ethan Steinberg, Ken Jung, Jason~A Fries, Conor~K Corbin, Stephen~R Pfohl, and Nigam~H Shah.
\newblock Language models are an effective representation learning technique for electronic health record data.
\newblock \emph{Journal of biomedical informatics}, 113:\penalty0 103637, 2021.

\bibitem[Sun et~al.(2023)Sun, Li, Li, Wu, Guo, Zhang, and Wang]{sun2023text}
Xiaofei Sun, Xiaoya Li, Jiwei Li, Fei Wu, Shangwei Guo, Tianwei Zhang, and Guoyin Wang.
\newblock Text classification via large language models.
\newblock \emph{arXiv preprint arXiv:2305.08377}, 2023.

\bibitem[Thirunavukarasu et~al.(2023)Thirunavukarasu, Ting, Elangovan, Gutierrez, Tan, and Ting]{thirunavukarasu2023large}
Arun~James Thirunavukarasu, Darren Shu~Jeng Ting, Kabilan Elangovan, Laura Gutierrez, Ting~Fang Tan, and Daniel Shu~Wei Ting.
\newblock Large language models in medicine.
\newblock \emph{Nature medicine}, pp.\  1--11, 2023.

\bibitem[Touvron et~al.(2023{\natexlab{a}})Touvron, Lavril, Izacard, Martinet, Lachaux, Lacroix, Rozi{\`e}re, Goyal, Hambro, Azhar, et~al.]{touvron2023llama1}
Hugo Touvron, Thibaut Lavril, Gautier Izacard, Xavier Martinet, Marie-Anne Lachaux, Timoth{\'e}e Lacroix, Baptiste Rozi{\`e}re, Naman Goyal, Eric Hambro, Faisal Azhar, et~al.
\newblock Llama: Open and efficient foundation language models.
\newblock \emph{arXiv preprint arXiv:2302.13971}, 2023{\natexlab{a}}.

\bibitem[Touvron et~al.(2023{\natexlab{b}})Touvron, Martin, Stone, Albert, Almahairi, Babaei, Bashlykov, Batra, Bhargava, Bhosale, et~al.]{touvron2023llama2}
Hugo Touvron, Louis Martin, Kevin Stone, Peter Albert, Amjad Almahairi, Yasmine Babaei, Nikolay Bashlykov, Soumya Batra, Prajjwal Bhargava, Shruti Bhosale, et~al.
\newblock Llama 2: Open foundation and fine-tuned chat models.
\newblock \emph{arXiv preprint arXiv:2307.09288}, 2023{\natexlab{b}}.

\bibitem[Vaswani et~al.(2017)Vaswani, Shazeer, Parmar, Uszkoreit, Jones, Gomez, Kaiser, and Polosukhin]{vaswani2017attention}
Ashish Vaswani, Noam Shazeer, Niki Parmar, Jakob Uszkoreit, Llion Jones, Aidan~N Gomez, {\L}ukasz Kaiser, and Illia Polosukhin.
\newblock Attention is all you need.
\newblock \emph{Advances in neural information processing systems}, 30, 2017.

\bibitem[Venigalla et~al.(2022)Venigalla, Frankle, and Carbin]{venigalla2022biomedlm}
A~Venigalla, J~Frankle, and M~Carbin.
\newblock Biomedlm: a domain-specific large language model for biomedical text.
\newblock \emph{MosaicML. Accessed: Dec}, 23\penalty0 (3):\penalty0 2, 2022.

\bibitem[Wasson et~al.(1985)Wasson, Sox, Neff, and Goldman]{wasson1985clinical}
John~H Wasson, Harold~C Sox, Raymond~K Neff, and Lee Goldman.
\newblock Clinical prediction rules: applications and methodological standards.
\newblock \emph{New England Journal of Medicine}, 313\penalty0 (13):\penalty0 793--799, 1985.

\bibitem[Wolf et~al.(2019)Wolf, Debut, Sanh, Chaumond, Delangue, Moi, Cistac, Rault, Louf, Funtowicz, et~al.]{wolf2019huggingface}
Thomas Wolf, Lysandre Debut, Victor Sanh, Julien Chaumond, Clement Delangue, Anthony Moi, Pierric Cistac, Tim Rault, R{\'e}mi Louf, Morgan Funtowicz, et~al.
\newblock Huggingface's transformers: State-of-the-art natural language processing.
\newblock \emph{arXiv preprint arXiv:1910.03771}, 2019.

\bibitem[Yang et~al.(2023{\natexlab{a}})Yang, Wu, Jiang, Lin, Gao, Danek, and Sun]{pyhealth2023yang}
Chaoqi Yang, Zhenbang Wu, Patrick Jiang, Zhen Lin, Junyi Gao, Benjamin Danek, and Jimeng Sun.
\newblock {PyHealth}: A deep learning toolkit for healthcare predictive modeling.
\newblock In \emph{Proceedings of the 27th ACM SIGKDD International Conference on Knowledge Discovery and Data Mining (KDD) 2023}, 2023{\natexlab{a}}.
\newblock URL \url{https://github.com/sunlabuiuc/PyHealth}.

\bibitem[Yang et~al.(2023{\natexlab{b}})Yang, Jin, Tang, Han, Feng, Jiang, Yin, and Hu]{yang2023harnessing}
Jingfeng Yang, Hongye Jin, Ruixiang Tang, Xiaotian Han, Qizhang Feng, Haoming Jiang, Bing Yin, and Xia Hu.
\newblock Harnessing the power of llms in practice: A survey on chatgpt and beyond.
\newblock \emph{arXiv preprint arXiv:2304.13712}, 2023{\natexlab{b}}.

\bibitem[Yang et~al.(2022)Yang, Chen, PourNejatian, Shin, Smith, Parisien, Compas, Martin, Costa, Flores, et~al.]{yang2022large}
Xi~Yang, Aokun Chen, Nima PourNejatian, Hoo~Chang Shin, Kaleb~E Smith, Christopher Parisien, Colin Compas, Cheryl Martin, Anthony~B Costa, Mona~G Flores, et~al.
\newblock A large language model for electronic health records.
\newblock \emph{NPJ Digital Medicine}, 5\penalty0 (1):\penalty0 194, 2022.

\bibitem[Zakka et~al.(2024)Zakka, Shad, Chaurasia, Dalal, Kim, Moor, Fong, Phillips, Alexander, Ashley, et~al.]{zakka2024almanac}
Cyril Zakka, Rohan Shad, Akash Chaurasia, Alex~R Dalal, Jennifer~L Kim, Michael Moor, Robyn Fong, Curran Phillips, Kevin Alexander, Euan Ashley, et~al.
\newblock Almanac—retrieval-augmented language models for clinical medicine.
\newblock \emph{NEJM AI}, 1\penalty0 (2):\penalty0 AIoa2300068, 2024.

\bibitem[Zhang et~al.(2021)Zhang, Gao, Ma, Wang, Wang, and Tang]{zhang2021grasp}
Chaohe Zhang, Xin Gao, Liantao Ma, Yasha Wang, Jiangtao Wang, and Wen Tang.
\newblock Grasp: generic framework for health status representation learning based on incorporating knowledge from similar patients.
\newblock In \emph{Proceedings of the AAAI conference on artificial intelligence}, volume~35, pp.\  715--723, 2021.

\bibitem[Zhao et~al.(2023)Zhao, Zhou, Li, Tang, Wang, Hou, Min, Zhang, Zhang, Dong, et~al.]{zhao2023survey}
Wayne~Xin Zhao, Kun Zhou, Junyi Li, Tianyi Tang, Xiaolei Wang, Yupeng Hou, Yingqian Min, Beichen Zhang, Junjie Zhang, Zican Dong, et~al.
\newblock A survey of large language models.
\newblock \emph{arXiv preprint arXiv:2303.18223}, 2023.

\end{thebibliography}
\bibliographystyle{iclr2024_conference}

\end{document}